\documentclass[conference,a4paper]{IEEEtran}
\IEEEoverridecommandlockouts
\usepackage{cite}
\usepackage{amsmath,amssymb,amsfonts}
\usepackage{algorithmic}
\usepackage{graphicx}
\usepackage{textcomp}
\usepackage{xcolor}
\usepackage{url}
\usepackage{booktabs, multirow, makecell, graphicx}
\def\BibTeX{{\rm B\kern-.05em{\sc i\kern-.025em b}\kern-.08em
    T\kern-.1667em\lower.7ex\hbox{E}\kern-.125emX}}
\begin{document}

\title{LEIQ-Assessor: Multi-dimensional Quality Assessment of Low-light Enhanced Images via Multi-task Learning
}

\author{Wei Sun\textsuperscript{1}, Yanwei Jiang\textsuperscript{2}, Dandan Zhu\textsuperscript{1}, Jinqiu Sang\textsuperscript{1},\\ Jikai Xu\textsuperscript{1}, Weixia Zhang\textsuperscript{2}, and Guangtao Zhai\textsuperscript{2}$^{\dag}$\thanks{$^{\dag}$Corresponding authors.}\thanks{This work was supported by the National Natural Science Foundation of China under Grants 62301316, 62377011, and 62225112.}\\
\textsuperscript{1}East China Normal University, \textsuperscript{2}Shanghai Jiao Tong University}


\maketitle

\begin{abstract}
Low-light image enhancement algorithms (LIEAs) aim to improve the visibility of images captured under poor illumination. However, the enhancement process often introduces artifacts such as noise amplification, color shift, structural damage, and over-exposure, which degrade the perceptual quality of the enhanced images. Therefore, a reliable image quality assessment (IQA) metric for evaluating enhancement effects is of great importance for both the development of LIEAs and their practical applications. In this paper, we present \textbf{LEIQ-Assessor}, a multi-dimensional quality assessment model for low-light image enhancement based on multi-task learning, developed for the QoMEX 2026 Grand Challenge on Low-light Enhanced Image Quality Assessment. Specifically, our method leverages a pre-trained SigLIP2 Vision Transformer as the backbone and simultaneously predicts the overall Mean Opinion Score (MOS) together with six perceptual sub-attributes: lightness, color fidelity, noise level, exposure quality, naturalness, and content recovery. By jointly optimizing these correlated objectives via the PLCC loss, the shared representation captures richer quality-aware features than its single-task counterpart. Experiments on the MLE benchmark demonstrate that LEIQ-Assessor significantly outperforms existing no-reference IQA models and hand-crafted quality descriptors. Our method achieved second place in the QoMEX 2026 Grand Challenge on Low-light Enhanced Image Quality Assessment. The code is available at \url{https://github.com/sunwei925/LEIQ-Assessor}.
\end{abstract}


\section{Introduction}
Images captured under low-light or backlight conditions frequently suffer from poor visibility, low contrast, and severe noise contamination, which not only degrade the quality of experience (QoE) but also impair the performance of downstream vision tasks such as object detection, semantic segmentation, and autonomous driving. To address this issue, a wide range of low-light image enhancement algorithms (LIEAs) have been proposed in recent years, spanning traditional histogram-based methods \cite{abdullah2007dynamic}, Retinex-based decomposition approaches \cite{fu2016weighted}, and deep-learning-driven frameworks \cite{jiang2021enlightengan}. While these methods have achieved remarkable progress in improving image visibility, the enhancement process itself is far from perfect. Artifacts including noise amplification, color shift, structural distortion, and overexposure are commonly observed in enhanced results, and such degradations can vary dramatically across different algorithms and scenes \cite{zhai2021perceptual}. This raises a critical question: \textit{how can we reliably and comprehensively assess the perceptual quality of low-light enhanced images?}

Recently, numerous general-purpose image quality assessment (IQA) metrics~\cite{mittal2012no,mittal2012making,sun2023blind,zhang2023blind,wang2023exploring,yang2022maniqa} and task-specific quality assessment metrics~\cite{sun2022deep,sun2020dynamic,sun2019mc360iqa,lu2022deep,sun2024analysis,sun2024dual,wang2024large,ge2025lmm,sun2025enhancing,sun2025empirical,sun2025compressedvqa,sun2024assessing,cao2026vqathinker,cao2026qualirag,cao2026generalizable,zhang2025benchmarking,sun2025efficient,zhu2025scandtm,zhu2024discrete,zhu2024mtcam} have been proposed, yet they perform poorly on low-light enhanced images, primarily due to the domain gap between domain-specific images and low-light enhanced images. Since most popular IQA methods are learning-based, they generalize poorly to this domain, as they are not designed to handle the unique distortion patterns introduced by LIEAs. To advance this field, Zhai~\textit{et al.}~\cite{zhai2021perceptual} constructed the first comprehensive low-light image enhancement quality (LIEQ) database, which includes 1,000 enhanced images derived from 100 low-light images using 10 LIEAs, and proposed a full-reference IQA metric named LIEQA that evaluates enhancement quality from four aspects—luminance enhancement, color rendition, noise evaluation, and structure preserving—to quantify the overall enhancement effectiveness. Further, Zhang~\textit{et al.}~\cite{zhang2021no} proposed a no-reference IQA method named NLIEE, which measures the quality of low-light enhanced images from four key perspectives: light enhancement, color comparison, noise measurement, and structure evaluation. Wang~\textit{et al.}~\cite{wang2025blind} presented a blind multimodal quality assessment (BMQA) framework for low-light images, which integrates image and text modalities to mimic how human visual perception leverages multiple sensory information when judging image quality, and constructed the multimodal low-light image quality (MLIQ) database containing 3,600 image-text pairs to validate its effectiveness. Although these efforts have significantly advanced the development of IQA for low-light enhanced images, they predominantly rely on a single scalar score to characterize complex and often conflicting quality attributes, which limits both the diagnostic value of the assessment and its utility for guiding algorithm development.

Therefore, recent efforts have begun to explore quality assessment specifically tailored to low-light image enhancement at a finer granularity. The QoMEX 2026 Grand Challenge on Low-light Enhanced Image Quality Assessment introduced the MLE benchmark~\cite{MLEDataset2026}, which provides subjective annotations not only for the overall Mean Opinion Score (MOS) but also for six fine-grained perceptual sub-attributes: lightness, color fidelity, noise level, exposure quality, naturalness, and content recovery. This benchmark opens up new opportunities for developing multi-dimensional IQA models that move beyond a single quality score and offer interpretable, attribute-level diagnostics.

In this paper, we present LEIQ-Assessor, a unified multi-dimensional quality assessment framework that jointly predicts the overall MOS and all six perceptual sub-attribute scores in a single forward pass. Our key insight is that these quality dimensions—such as lightness, color fidelity, and noise level—are not independent but exhibit strong inter-attribute correlations, and modeling them jointly enables the network to exploit shared perceptual cues that benefit all predictions simultaneously. Specifically, we adopt a pre-trained SigLIP2~\cite{tschannen2025siglip} Vision Transformer (ViT) as the feature extraction backbone, whose sigmoid-based vision–language contrastive pre-training yields representations that are both semantically rich and perceptually discriminative. Upon this foundation, we introduce a multi-head prediction module with dimension-specific lightweight regression heads, jointly optimized via a composite Pearson Linear Correlation Coefficient (PLCC) loss that directly maximizes the linear agreement between predictions and ground-truth annotations across all quality dimensions. This multi-task formulation drives the shared backbone to distill quality-aware representations that capture cross-attribute dependencies, effectively enabling knowledge transfer among correlated dimensions and yielding stronger generalization than independently trained single-task counterparts.

The main contributions of this work are summarized as follows:

\begin{itemize} \item We propose LEIQ-Assessor, a multi-task learning framework that jointly predicts the overall perceptual quality score and six fine-grained dimensional scores for low-light enhanced images, providing both an accurate global assessment and interpretable attribute-level diagnostics. \item We demonstrate that a pre-trained SigLIP2 Vision Transformer, when combined with PLCC-based multi-task optimization, serves as a highly effective backbone for capturing quality-aware features specific to enhancement artifacts. \item Extensive experiments on the MLE benchmark show that LEIQ-Assessor significantly outperforms existing no-reference IQA models. What's more, our method achieved second place in the QoMEX 2026 Grand Challenge on Low-light Enhanced Image Quality Assessment, further validating its effectiveness. \end{itemize}

\section{Method}

\subsection{Overview}

\begin{figure}[t]
    \centering
    \includegraphics[width=0.42\textwidth]{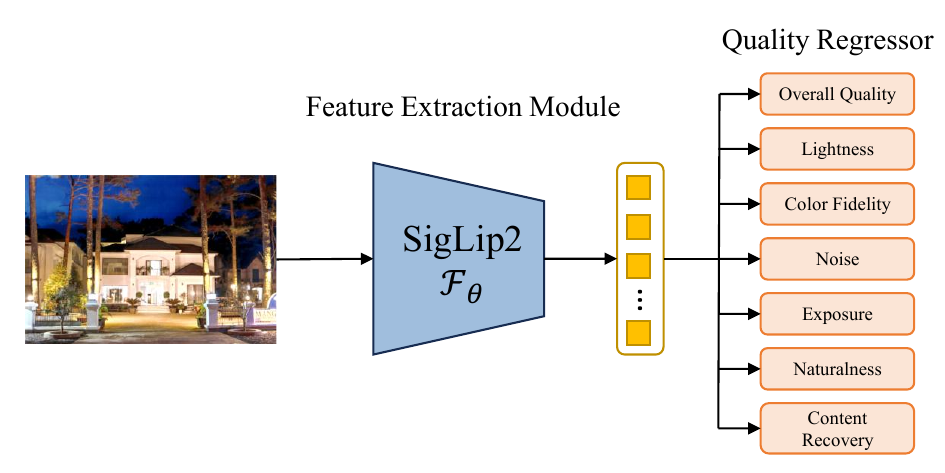}  
    \caption{Overall architecture of the proposed LEIQ-Assessor. A low-light enhanced image is fed into the pre-trained SigLIP2 Vision Transformer backbone $\mathcal{F}_\theta$ to extract a shared feature representation, which is then mapped by seven dimension-specific regression heads to simultaneously predict the overall quality score (MOS) and six dimensional scores: lightness, color fidelity, noise level, exposure quality, naturalness, and content recovery.}
    \label{fig:framework}
\end{figure}

Given a low-light enhanced image $\mathbf{I} \in \mathbb{R}^{H \times W \times 3}$, our goal is to simultaneously predict the overall Mean Opinion Score (MOS) and six fine-grained dimensional scores, including lightness, color fidelity, noise level, exposure quality, naturalness, and content recovery. We denote the set of all quality dimensions as $\mathcal{T} = \{\text{mos},\, \text{light},\, \text{color},\, \text{noise},\, \text{exposure},\, \text{nature},\, \text{content\_recovery}\}$, resulting in a seven-dimensional prediction vector. As illustrated in Fig.~\ref{fig:framework}, LEIQ-Assessor follows a simple yet effective encoder--regressor paradigm: a shared Vision Transformer (ViT) backbone first extracts a rich feature representation from the input image, after which a set of dimension-specific regression heads independently map this shared representation to the corresponding quality scores. All components are trained end-to-end with a composite loss that directly maximizes the Pearson Linear Correlation Coefficient (PLCC) between predictions and ground-truth annotations across all quality dimensions.

\subsection{Feature Extraction Module}

\begin{table*}[t]
  \centering
  \caption{Comparison of IQA methods on the MLE dataset. Mean SRCC / PLCC over 10 random splits are reported. \textbf{Bold} indicates the best result.}
  \label{tab:multitask_results}
  \setlength{\tabcolsep}{7pt}
  \small
  \begin{tabular}{l ccccccc}
    \toprule
    Task & BRISQUE & NIQE & MANIQA & StairIQA & CLIP-IQA+ & LIQE & LEIQ-Assessor \\
    \midrule
    MOS          & 0.365 / 0.402 & 0.365 / 0.406 & 0.274 / 0.315 & 0.425 / 0.442 & 0.489 / 0.501 & 0.595 / 0.603 & \textbf{0.886} / \textbf{0.892} \\
    Light        & 0.117 / 0.189 & 0.286 / 0.408 & 0.178 / 0.262 & 0.341 / 0.412 & 0.173 / 0.256 & 0.339 / 0.382 & \textbf{0.796} / \textbf{0.863} \\
    Color        & 0.279 / 0.329 & 0.335 / 0.383 & 0.231 / 0.271 & 0.384 / 0.374 & 0.412 / 0.434 & 0.515 / 0.526 & \textbf{0.821} / \textbf{0.830} \\
    Noise        & 0.259 / 0.301 & 0.297 / 0.335 & 0.392 / 0.412 & 0.458 / 0.460 & 0.552 / 0.571 & 0.655 / 0.662 & \textbf{0.831} / \textbf{0.850} \\
    Exposure     & 0.336 / 0.382 & 0.195 / 0.217 & 0.160 / 0.128 & 0.282 / 0.278 & 0.389 / 0.363 & 0.439 / 0.393 & \textbf{0.817} / \textbf{0.881} \\
    Nature       & 0.369 / 0.409 & 0.318 / 0.342 & 0.282 / 0.285 & 0.401 / 0.423 & 0.480 / 0.483 & 0.586 / 0.586 & \textbf{0.888} / \textbf{0.906} \\
    Content Rec. & 0.306 / 0.351 & 0.380 / 0.437 & 0.311 / 0.330 & 0.478 / 0.474 & 0.488 / 0.498 & 0.610 / 0.611 & \textbf{0.863} / \textbf{0.870} \\
    \midrule
    Average      & 0.290 / 0.338 & 0.311 / 0.361 & 0.261 / 0.286 & 0.396 / 0.409 & 0.426 / 0.444 & 0.534 / 0.538 & \textbf{0.843} / \textbf{0.870} \\
    \bottomrule
  \end{tabular}
\end{table*}

We adopt a pre-trained SigLIP2 Vision Transformer~\cite{tschannen2025siglip} as the shared feature extraction backbone. Compared with conventional CLIP-based encoders, SigLIP2 replaces the softmax-based contrastive loss with a pairwise sigmoid loss during vision--language pre-training, which encourages the learned representations to encode semantically fine-grained attributes rather than only coarse categorical distinctions---a property particularly beneficial for quality assessment where subtle perceptual differences must be captured. Moreover, SigLIP2 is pre-trained on a massive web-scale dataset with diverse visual content, endowing the backbone with strong cross-domain generalization capability. Given an input image $\mathbf{I}$, we feed it into the SigLIP2 ViT encoder $\mathcal{F}_{\theta}$ to obtain a global feature representation:
\begin{equation}
    \mathbf{f} = \mathcal{F}_{\theta}(\mathbf{I}) \in \mathbb{R}^{D},
    \label{eq:feature_extraction}
\end{equation}
where $D$ denotes the output feature dimension of the ViT encoder, and $\theta$ represents the pre-trained backbone parameters that are fine-tuned during training. The feature vector $\mathbf{f}$ encodes rich semantic and perceptual information and serves as the shared representation fed into all downstream dimension-specific regression heads.

\subsection{Multi-Head Quality Regressor}

On top of the shared feature representation $\mathbf{f}$, we attach seven lightweight, dimension-specific regression heads, one for each quality dimension $t \in \mathcal{T}$. Each head is implemented as a compact multi-layer perceptron (MLP) consisting of two successive linear projections that map the shared feature to a scalar prediction:
\begin{equation}
    \hat{y}_t = \mathcal{H}_t(\mathbf{f}) = \mathbf{W}_t^{(2)} \left(\mathbf{W}_t^{(1)} \mathbf{f}\right), \quad t \in \mathcal{T},
    \label{eq:head}
\end{equation}
where $\mathbf{W}_t^{(1)} \in \mathbb{R}^{d_h \times D}$ and $\mathbf{W}_t^{(2)} \in \mathbb{R}^{1 \times d_h}$ are the learnable parameters of head $t$, and $d_h$ is the hidden dimension. The use of separate heads allows each dimension to specialize independently, while the shared backbone ensures cross-attribute knowledge sharing with negligible parameter overhead.

\subsection{PLCC-Based Multi-Task Loss}

A natural choice for regression losses, such as Mean Squared Error (MSE) or Mean Absolute Error (MAE), penalizes point-wise prediction errors but does not directly optimize the correlation between predicted and ground-truth score distributions. In subjective quality assessment, however, the ranking consistency and linear agreement between predictions and human annotations are the primary evaluation criteria, as reflected by the Spearman Rank-Order Correlation Coefficient (SRCC) and PLCC metrics widely adopted in the IQA community. To bridge this gap, we employ a differentiable PLCC loss that directly maximizes the Pearson correlation within each training mini-batch. The total training loss is the unweighted sum of the per-dimension PLCC losses across all seven quality dimensions:
\begin{equation}
    \mathcal{L} = \sum_{t \in \mathcal{T}} \mathcal{L}_t^{\mathrm{PLCC}}.
    \label{eq:total_loss}
\end{equation}

\section{Experimental Validation}

\subsection{Experimental Protocol}
We train and evaluate our model on the MLE dataset provided by the QoMEX 2026 Grand Challenge on Low-light Enhanced Image Quality Assessment~\cite{MLEDataset2026}, which contains 800 low-light enhanced images annotated with seven-dimensional quality labels. The dataset is randomly split into training and test sets with a ratio of 8:2, and we report results averaged over 10 random splits to account for variance.

The entire framework is trained end-to-end using the Adam optimizer with an initial learning rate of $1 \times 10^{-5}$ and a weight decay of $1 \times 10^{-7}$. Training proceeds for 30 epochs with a mini-batch size of 8. We compare our method against several representative IQA models, including BRISQUE~\cite{mittal2012no}, NIQE~\cite{mittal2012making}, MANIQA~\cite{yang2022maniqa}, StairIQA~\cite{sun2023blind}, CLIP-IQA+~\cite{wang2023exploring} and LIQE~\cite{zhang2023blind}. The prediction performance is evaluated using two widely adopted criteria: PLCC and SRCC.

\subsection{Experimental Results}
We present the experimental results in Table~\ref{tab:multitask_results}. As shown in Table~\ref{tab:multitask_results}, both traditional handcrafted IQA methods and recent DNN-based methods perform poorly on the MLE dataset. This reveals a significant gap between low-light enhanced image quality assessment and general image quality assessment, suggesting that existing IQA models fail to capture the unique distortion characteristics introduced by low-light enhancement algorithms. It is therefore necessary to construct low-light enhancement oriented IQA datasets and train expert IQA models specifically tailored for such images. Our method, LEIQ-Assessor, leverages the powerful feature extraction capability of SigLIP2 and, after fine-tuning, achieves the best performance across all quality dimensions. Furthermore, it attained second place in the QoMEX 2026 Grand Challenge on Low-light Enhanced Image Quality Assessment, demonstrating its effectiveness.

\section{Conclusion}
In this paper, we have presented LEIQ-Assessor, a multi-task learning framework that jointly predicts the overall MOS and six dimensional scores for low-light enhanced images by leveraging a pre-trained SigLIP2 Vision Transformer backbone and a composite PLCC loss. This multi-task formulation drives the shared backbone to distill quality-aware representations that capture cross-attribute dependencies, effectively enabling knowledge transfer among correlated dimensions and yielding stronger generalization than independently trained single-task counterparts. Extensive experiments on the MLE benchmark demonstrate that our method significantly outperforms existing no-reference IQA models across all quality dimensions, achieving second place in the QoMEX 2026 Grand Challenge on Low-light Enhanced Image Quality Assessment.

\bibliographystyle{IEEEbib}
\bibliography{reference}

\end{document}